\documentclass[runningheads]{llncs}
\usepackage{graphicx}
\usepackage{amsmath,amssymb} 
\usepackage{color}
\usepackage[width=122mm,left=12mm,paperwidth=146mm,height=193mm,top=12mm,paperheight=217mm]{geometry}
\usepackage[ruled,vlined]{algorithm2e}
\usepackage{subfigure}
\begin{document}
\pagestyle{headings}
\mainmatter

\title{Embedding Deep Metric for Person Re-identification: A Study Against Large Variations} 

\titlerunning{Embedding Deep Metric for Person Re-identification: A Study Against Large Variations}

\authorrunning{Hailin Shi \emph{et al.}}

\author{Hailin Shi$^{1,2}$,
Yang Yang$^{1,2}$,
Xiangyu Zhu$^{1,2}$,
Shengcai Liao$^{1,2}$,
Zhen Lei$^{1,2}$\thanks{Corresponding Author.},
Weishi Zheng$^3$,
Stan Z. Li$^{1,2}$}

\institute{$^1$Center for Biometrics and Security Research \& National Laboratory of Pattern Recognition, Institute of Automation,
Chinese Academy of Sciences\\
$^2$University of Chinese Academy of Sciences\\
$^3$School of Data and Computer Science, Sun Yat-sen University \\
\email{\{hailin.shi,yang.yang,xiangyu.zhu,scliao,zlei,szli\}@nlpr.ia.ac.cn}}


\maketitle

\begin{abstract}
   Person re-identification is challenging due to the large variations of pose, illumination, occlusion and camera view.
   Owing to these variations, the pedestrian data is distributed as highly-curved manifolds in the feature space, despite the current convolutional neural networks (CNN)'s capability of feature extraction.
   However, the distribution is unknown, so it is difficult to use the geodesic distance when comparing two samples.
   In practice, the current deep embedding methods use the Euclidean distance for the training and test.
   On the other hand, the manifold learning methods suggest to use the Euclidean distance in the local range, combining with the graphical relationship between samples, for approximating the geodesic distance.
   From this point of view, selecting suitable positive (\emph{i.e.} intra-class) training samples within a local range is critical for training the CNN embedding, especially when the data has large intra-class variations.
   In this paper, we propose a novel moderate positive sample mining method to train robust CNN for person re-identification, dealing with the problem of large variation.
   In addition, we improve the learning by a metric weight constraint, so that the learned metric has a better generalization ability.
   Experiments show that these two strategies are effective in learning robust deep metrics for person re-identification, and accordingly our deep model significantly outperforms the state-of-the-art methods on several benchmarks of person re-identification.
   Therefore, the study presented in this paper may be useful in inspiring new designs of deep models for person re-identification.
\end{abstract}

\section{Introduction}
\label{section_Introduction}

Given a set of pedestrian images, person re-identification aims to identify the probe image that generally captured by different cameras. Nowadays, person re-identification becomes increasingly important for surveillance and security system, \emph{e.g.} replacing manual video screening and other heavy loads.
Person re-identification is a challenging task due to large variations of body pose, lighting, view angles, scenarios across time and cameras.

The framework of existing methods usually consists of two parts: (1) extracting discriminative features from pedestrian images; (2) computing the distance of samples by feature comparison.
There are many works focus on these two aspects.
The traditional methods work at improving suitable hand-crafted features~\cite{yang2014salient,zhao2014learning}, or good metric for comparison~\cite{paisitkriangkrai2015learning,koestinger2012large,li2013locally,li2013learning,martinel2014saliency,zhao2013person,zhang2014prism,zhang2015group}, or both of them~\cite{khamis2014joint,liao2015person,xiong2014person,zhang2014novel}. The first aspect considers to find features that are robust to challenging factors (lighting, pose \emph{etc.}) while preserving the identity information. The second aspect comes to the metric learning problem which generally minimizes the intra-class distance while maximizing the inter-class distance.

More recently, the deep learning methods gradually gain the popularity in person re-identification.
The re-identification methods by deep learning~\cite{ahmed2015improved,ding2015deep,li2014deepreid,yi2014deep} incorporate the two above-mentioned aspects (feature extraction and metric learning) of person re-identification into an integrated framework.
The feature extraction and the metric learning are fulfilled respectively by two components in a deep neural network: the CNN part which extracts features from images, and the following metric learning part which compares the features with the metric.
The FPNN~\cite{li2014deepreid} algorithm introduced a patch matching layer for the CNN part for the first time. Ahmed \emph{et al.}~\cite{ahmed2015improved} proposed an improved deep learning architecture (IDLA) with cross-input neighborhood differences and patch summary features. These two methods are both dedicated to improve the CNN architecture.
Their purpose is to evaluate the pair similarity early in the CNN stage, so that it could make use of spatial correspondence of feature maps.
As for the metric learning part, DML~\cite{yi2014deep} adopted the cosine similarity and Binomial deviance.
DeepFeature~\cite{ding2015deep} adopted the Euclidean distance and triplet loss.
Some others~\cite{ahmed2015improved,li2014deepreid} used the logistic loss to directly form a binary classification problem of whether the input image pair belongs to the same identity.

The following are our contributions.
\begin{itemize}\itemsep=-1pt
    \item   For training the CNN, the hard negative mining strategy has been used in ~\cite{ahmed2015improved,parkhi2015deep,schroff2015facenet}.
    Considering the large intra-class variations in pedestrian data, we argue that, in person re-identification,
    the positive training pairs should also be sampled carefully since the pedestrian data is distributed as the manifold that are highly curved in the feature space.
    As argued in some manifold learning methods~\cite{tenenbaum2000global,roweis2000nonlinear,belkin2003laplacian}, it is effective to use the local Euclidean distance, combining with the graphical relationship between samples, to approximate the geodesic distance.
    Thus, selecting the moderate positive pairs in the local range is critical for training the network.
    This is an important issue but has been seldom noticed.
    In this paper, we propose a new training strategy, named \textbf{moderate positive mining}\footnote{The source codes is available at http://www.cbsr.ia.ac.cn/users/hailinshi.
}, to adaptively search the moderate positives for training.
    This novel training method significantly improves the identification accuracy.
    \item   In addition, we improve the network by the \textbf{weight constraint} for the metric layers.
    The weight constraint regularizes the metric learning part and alleviates the over-fitting problem.
\end{itemize}

\section{Related Work}
\label{section_Related_work}

\subsubsection{Positive Sample Mining.}
The hard negative mining strategy~\cite{schroff2015facenet} has been used for face recognition.
In person re-identification, IDLA~\cite{ahmed2015improved} also adopted hard negative mining for the training.
By forcing the model to focus on the hard negatives near the decision boundary, hard negative mining improves the training efficiency and the model performance.
In this paper, we find that how to select moderate positive samples is also an essential issue for learning person re-identification model.
The moderate positives are as critical as hard negatives for training the network, especially when the data has large intra-class variations.
However, there are barely any previous attempt in this aspect for learning the deep embedding.
In our approach, we propose the novel strategy of moderate positive mining to address the problem.
We sample the moderate positives for training, and avoid using the hard ones from extreme intra-class variations of pedestrian data.
We empirically find that this strategy effectively improves the identification accuracy (see Section~\ref{section_Analysis of Moderate Positive Mining}).

\subsubsection{Weight Constraint for Metric Learning.}
A commonly used metric by deep learning methods is the Euclidean distance~\cite{ding2015deep,schroff2015facenet,parkhi2015deep}.
However, the Euclidean distance is sensitive to the scale, and is blind to the correlation across dimensions.
In practice, we cannot guarantee the CNN-learned features have similar scales and the de-correlation across dimensions.
Therefore, using the Mahalanobis distance is a better choice for multivariate metric~\cite{manly2004multivariate}.
In the area of face recognition, DDML~\cite{hu2014discriminative} implemented the Mahalanobis metric in their network, but without any constraint.
Our metric is learned in a similar way and improved by the proposed weight constraint which helps to gain a better generalization ability.

\section{Proposed Method}
\label{section_Proposed Method}

In this section, we firstly introduce the moderate positive mining method.
Then, we revisit DDML and introduce the weight constraint.

\subsection{Moderate Positive Mining}
\label{section_Moderate Positive Mining}

\subsubsection{Large Intra-class Variations}
There are many factors lead to the large intra-class variations in pedestrian data, such as illumination, background, misalignment, occlusion, co-occurrence of people, appearance changing, \emph{etc.}
Many of them are specific with pedestrian data.
Fig.~\ref{fig_hard-positives}(a) shows some hard positive cases in the data set of CUHK03~\cite{li2014deepreid}.
Some of them are even difficult for human to recognize.

Although CNN has a strong ability to extract features, pedestrian data follows the very irregular distribution in the feature space due to the large variations, such as the example of highly-curved manifold illustrated in Fig.~\ref{fig_hard-positives}(b).
This is reflected by the fact the state-of-the-art performances on several person re-identification benchmarks are relatively poor comparing with the human face recognition task which is easier due to less intra-class variations.

\subsubsection{Moderate Positive Mining Method}
\begin{figure*}[!htbp]
  \centering

  \subfigure[]{
  \includegraphics[width=0.32\textwidth]{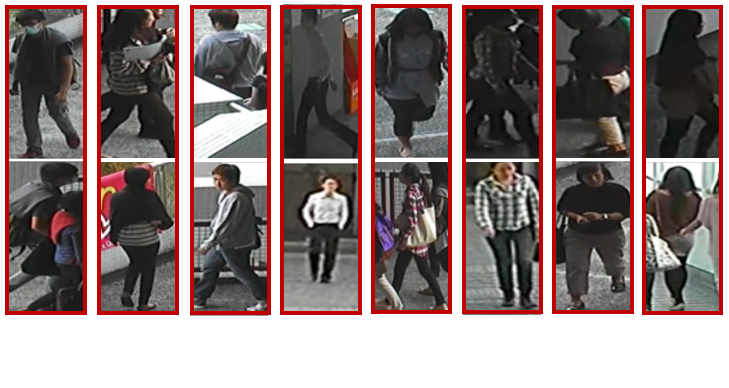}}
  \subfigure[]{
  \includegraphics[width=0.30\textwidth]{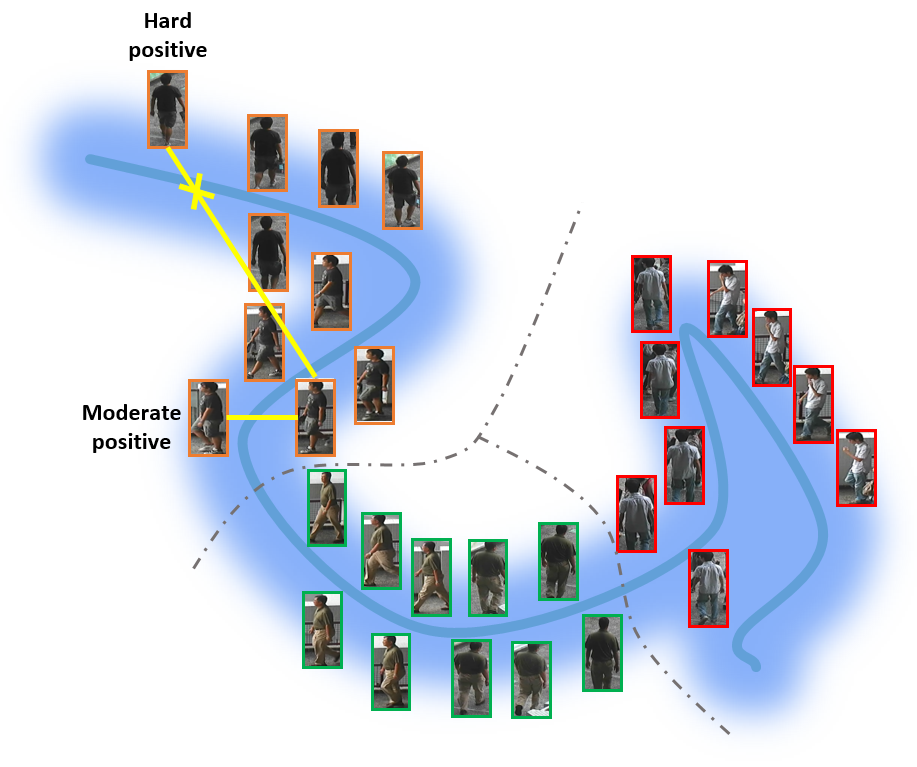}}
  \subfigure[]{
  \includegraphics[width=0.30\textwidth]{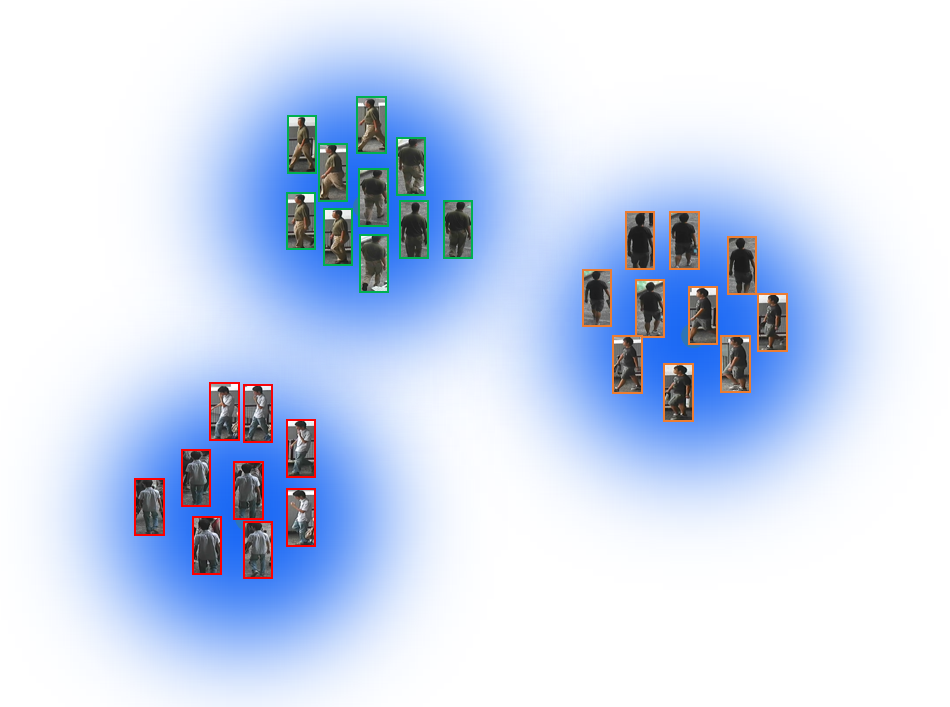}}

  \caption{(a) Some hard positive cases in CUHK03. (b) Illustration of the highly-curved manifold of 3 identities. (c) Gaussian distribution is suitable to perform Mahalanobis metric. Best viewed in color.}
  \label{fig_hard-positives}
\end{figure*}

Considering the distribution in Fig.~\ref{fig_hard-positives}(b) is unknown, it is difficult to apply the geodesic distance for comparing two samples. The usual way is to use the Mahalanobis distance (or the special case Euclidean)~\cite{ding2015deep,schroff2015facenet,hu2014discriminative} which is a suitable metric in the ideal condition (Fig.~\ref{fig_hard-positives}(c)).

On the other hand, the manifold learning methods~\cite{tenenbaum2000global,roweis2000nonlinear,belkin2003laplacian} suggest to use the Euclidean distance (or heat kernel) in the local range, combining with the graphical relationship between samples, for approximating the geodesic distance.
This is a feasible way to minimize the intra-class variance along the manifold for the supervised learning.
However, when training the deep CNN with contrastive or triplet loss for embedding, the existing deep embedding methods use the Euclidean distance undiscriminatingly with all the positive samples.

Here, we argue that selecting positive samples in the local range (pairing by the yellow line in Fig.~\ref{fig_hard-positives}(b)) is critical for training the network; training with the positive samples of large distance (the yellow line with cross) may distort the manifold and harm the manifold learning.

The basic idea is that we reduce the intra-class variance while preserving the intrinsic graphical structure of pedestrian data via mining the moderate positive pairs in the local range.

We introduce the moderate positive mining method as follows: we select the moderate positive pairs in the range of the same subject at one time.
For example, suppose a subject having 6 images, of which 3 from a camera and 3 from another.
We can totally match 9 positive pairs from this subject.
If we use the easiest positive pair of the nine, the convergence will be very slow;
if we use the hardest, the learning will be damaged.
Thus, we pick the moderate positive pairs that are between the two extreme cases.

Given two sets of pedestrian images $\mathcal{I}_\mathbf{1}$ and $\mathcal{I}_\mathbf{2}$  come from two disjoint cameras.
Denote $\mathbf{I}_1 \in \mathcal{I}_\mathbf{1}$ and $\mathbf{I}_2^p \in \mathcal{I}_\mathbf{2}$ as a positive pair (from the same identity),
and $\mathbf{I}_1 \in \mathcal{I}_\mathbf{1}$ and $\mathbf{I}_2^n \in \mathcal{I}_\mathbf{2}$ as a negative pair (from different identities).
Denote $\mathbf{\Psi} (\cdot)$ as the CNN, $ d(\cdot, \cdot) $ is the Mahalanobis or Euclidean distance.
The mining method is described as follows:

\begin{algorithm}[h] \label{algorithm-mining}
	\SetAlgoLined \caption{\small Moderate Positive Mining}
	\begin{small}
        \KwIn{randomly select an anchor sample $\mathbf{I}_1$, its positive samples $\{\mathbf{I}_2^{p_1}, \dots, \mathbf{I}_2^{p_k}\}$
        and negative samples $\{\mathbf{I}_2^{n_1}, \dots, \mathbf{I}_2^{n_k}\}$ to form a mini-batch.}

        \textbf{Step 1}     Input the images into the network for obtaining the features, and compute their distances
        $ \{ d(\mathbf{\Psi}(\mathbf{I}_1), \mathbf{\Psi}(\mathbf{I}_2^{p_1})), \dots, d(\mathbf{\Psi}(\mathbf{I}_1), \mathbf{\Psi}(\mathbf{I}_2^{p_k})) \}$
        and
        $ \{ d(\mathbf{\Psi}(\mathbf{I}_1), \mathbf{\Psi}(\mathbf{I}_2^{n_1})), \dots, d(\mathbf{\Psi}(\mathbf{I}_1), \mathbf{\Psi}(\mathbf{I}_2^{n_k})) \}$;

        \textbf{Step 2}     mine the hardest negative sample $$\hat{\mathbf{I}}_2^n = argmin_{j=1 \dots k}\{ d(\mathbf{\Psi}(\mathbf{I}_1), \mathbf{\Psi}(\mathbf{I}_2^{n_j}))  \};$$

        \textbf{Step 3}     from the positive samples, choose those $ \tilde{\mathbf{I}}_2^{p_m} $ satisfying
        $$ d(\mathbf{\Psi}(\mathbf{I}_1), \mathbf{\Psi}(\tilde{\mathbf{I}}_2^{p_m})) \le d(\mathbf{\Psi}(\mathbf{I}_1), \mathbf{\Psi}(\hat{\mathbf{I}}_2^n)); $$

        \textbf{Step 4}     mine the hardest one among these chosen positives as our moderate positive sample

        $$ \hat{\mathbf{I}}_2^p = argmax_{\tilde{\mathbf{I}}_2^{p_m}} \{ d(\mathbf{\Psi}(\mathbf{I}_1), \mathbf{\Psi}(\tilde{\mathbf{I}}_2^{p_m})) \}. $$
        If none of the positives satisfies the condition in \textbf{Step 3}, choose the positive with the smallest distance as the moderate positive sample.

        \KwOut{The moderate positive sample $\hat{\mathbf{I}}_2^p$.}

	\end{small}
	\label{alg:mpm}
\end{algorithm}

Firstly, we randomly select an anchor sample and its positive samples and negative samples (with equal number) to form a mini-batch;
then, we mine the hardest negative sample, and choose the positive samples that have smaller distances than the hardest negative;
finally, we mine the hardest one among these chosen positives as our moderate positive sample.
The reason to do so is that we define the ``moderate positive'' adaptively within each subject while their hard negatives are also involved in case the positives are too easy or too hard to be mined.

\begin{figure}[!htb]
  \centering
  \includegraphics[width=0.46\textwidth]{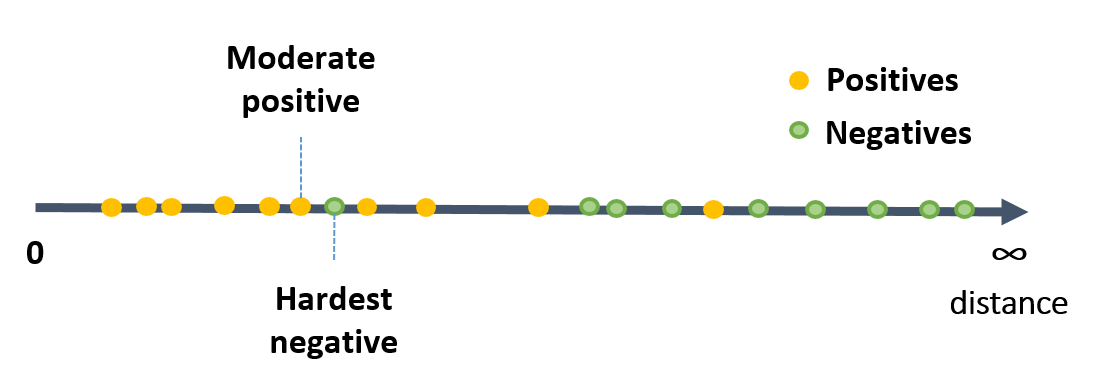}
  \caption{An example of the moderate positive mining in learning.}
  \label{fig_mining_diagram}
\end{figure}

An example is given in Fig.~\ref{fig_mining_diagram}.
In the experiments, this dynamic mining strategy improves the performance significantly, and shows good stability since all the positives are considered in each subject and the data is augmented by random translation.

\subsection{Weight Constraint for Deep Metric Learning}
\label{section_Constrained Deep Metric Learning}
Once the CNN extract the features from a pair of images, the metric layers are performed subsequently to calculate the distance, as shown in Fig.~\ref{fig_metric_learning}. The metric learning layer is like the structure proposed in DDML~\cite{hu2014discriminative}, and its learning is improved via a weight constraint.

Recalling the two sets of pedestrian images $\mathcal{I}_\mathbf{1}$ and $\mathcal{I}_\mathbf{2}$ mentioned above, denote $\mathcal{X}_\mathbf{1}$ and $\mathcal{X}_\mathbf{2}$ are the corresponding feature sets extracted by the CNN.
$\mathbf{x}_1 = \mathbf{\Psi}(\mathbf{I}_1)$, $\mathbf{x}_2^p = \mathbf{\Psi}(\mathbf{I}_2^p)$ and $\mathbf{x}_2^n = \mathbf{\Psi}(\mathbf{I}_2^n)$ are the corresponding features of the anchor, positive and negative samples.

\subsubsection{Revisiting DDML}

\begin{figure}[!htb]
  \centering
  \includegraphics[width=0.36\textwidth]{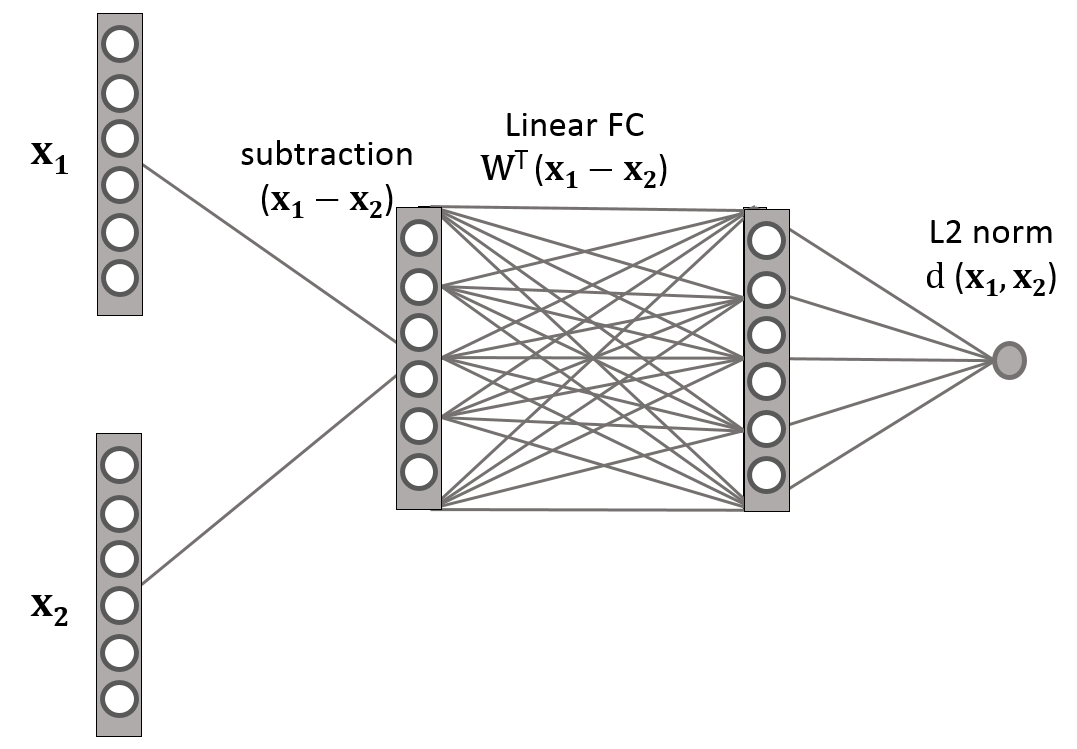}
  \caption{The metric learning layers compute the distance of two samples. $\mathbf{x}_1$ and $\mathbf{x}_2$ are the feature vectors extracted by the CNN from the images. The weight $\textbf{W}$ is regularized by the proposed constraint in the learning.}
  \label{fig_metric_learning}
\end{figure}

The Mahalanobis distance is formulated as
\begin{equation}\label{M-dist}
    d(\mathbf{x}_1, \mathbf{x}_2) = \sqrt{ (\mathbf{x}_1 - \mathbf{x}_2)^T \textbf{M} (\mathbf{x}_1 - \mathbf{x}_2) },
\end{equation}
where $\mathbf{x}_2 \in \{\mathbf{x}_2^p,\mathbf{x}_2^n\}$, $\textbf{M}$ is a symmetric positive semi-definite matrix.
Learning $\textbf{M}$ under the constraint of positive semi-definite is difficult. We make use of its decomposition $\textbf{M} = \textbf{W}\textbf{W}^T$. Learning $\textbf{W}$ is much easier, and $\textbf{W}\textbf{W}^T$ is always positive semi-definite.
We develop the distance as follows
\begin{align}\label{M-dist-dev-2}
    d(\mathbf{x}_1, \mathbf{x}_2)  & = \sqrt{ (\mathbf{x}_1 - \mathbf{x}_2)^T \textbf{W}\textbf{W}^T (\mathbf{x}_1 - \mathbf{x}_2) }    \nonumber   \\
    & = \sqrt{ (\textbf{W}^T(\mathbf{x}_1 - \mathbf{x}_2))^T(\textbf{W}^T(\mathbf{x}_1 - \mathbf{x}_2)) }            \nonumber   \\
    & = \|\textbf{W}^T(\mathbf{x}_1 - \mathbf{x}_2)\|_2.
\end{align}
The inner product $\textbf{W}^T(\mathbf{x}_1 - \mathbf{x}_2)$ can be implemented by a linear fully-connected (FC) layer in which the weight matrix is defined by $\textbf{W}^T$.
The output of the FC layer is calculated by
\begin{equation}\label{FC_compute}
    \mathbf{y} = f(\textbf{W}^T\mathbf{x} + \mathbf{b}),
\end{equation}
where $\mathbf{b}$ is the bias term.
The identity function is used as the activation $f(\cdot)$ for the linear FC layer.
As shown in Fig.~\ref{fig_metric_learning}, the feature vectors $\mathbf{x}_1$ and $\mathbf{x}_2$ are fed into the subtraction layer.
Then, the difference is transformed by the linear FC layer with the weight matrix $\textbf{W}^T$.
For the symmetry of the distance, we fix the bias term $\mathbf{b}$ to zero throughout the training and test.
Finally, the L2 norm is computed as the output distance $d(\mathbf{x}_1, \mathbf{x}_2)$.
This structure remains equivalent when switching the position of the subtraction layer and the FC layer.

\subsubsection{Weight Constraint}
The objective is to minimize the intra-class distance and maximize the inter-class distance.
The training loss is defined as
\begin{equation}\label{Loss}
    \mathbf{\emph{L}} = d(\mathbf{\Psi}(\mathbf{I}_1), \mathbf{\Psi}(\mathbf{I}_2^p)) +
    [m - d(\mathbf{\Psi}(\mathbf{I}_1), \mathbf{\Psi}(\mathbf{I}_2^n))]_{+},
\end{equation}
where $\mathbf{I}_1$, $\mathbf{I}_2^p$ and $\mathbf{I}_2^n$ are the input images corresponding to the features $\mathbf{x}_1$, $\mathbf{x}_2^p$ and $\mathbf{x}_2^n$, and $m$ is the margin which is set to 2 in the implementation.
In each time of the forward propagation, either the first term or the second term of Eq.~\ref{Loss} is computed.
Then the loss is obtained by combining the two terms, and we compute the gradient.

Compared with the Mahalanobis distance, the Euclidean distance has less discriminability but better generalization ability, because it does not take account of the scales and the correlation across dimensions~\cite{manly2004multivariate}.
Here, we impose a constraint that keep the matrix $\textbf{M}$ having large values at the diagonal and small entries elsewhere, so we can achieve a balance between the unconstrained Mahalanobis distance and the Euclidean distance.
The constraint is formulated as the Frobenius norm of the difference between $\textbf{W}\textbf{W}^T$ and identity matrix $\mathbf{I}$,
\begin{align}\label{Loss-st}
    \mathbf{\emph{L}} = d(\mathbf{\Psi}(\mathbf{I}_1), \mathbf{\Psi}(\mathbf{I}_2^p)) + [m - d(\mathbf{\Psi}(\mathbf{I}_1), \mathbf{\Psi}(\mathbf{I}_2^n))]_{+}      \nonumber   \\
    s.t. \quad \|\textbf{W}\textbf{W}^T - \mathbf{I}\|^2_F \leq \emph{C},
\end{align}
where $\emph{C}$ is a constant. We further combine the constraint into the loss function as a regularization term:
\begin{equation}\label{Loss-lambda}
    \hat{ \mathbf{\emph{L}} } = \mathbf{\emph{L}} + \frac{\lambda}{2} \|\textbf{W}\textbf{W}^T - \mathbf{I}\|^2_F,
\end{equation}
where $\lambda$ is the relative weight of regularization, $\hat{ \mathbf{\emph{L}} }$ is the new loss function.
For updating the weight matrix $\textbf{W}$, the gradient w.r.t $\textbf{W}$ is computed by
\begin{equation}\label{Loss-gradient}
    \frac{\partial \hat{ \mathbf{\emph{L}} }}{\partial \textbf{W}} =
    \frac{\partial \mathbf{\emph{L}}}{\partial \textbf{W}} + \lambda (\textbf{W}\textbf{W}^T - \mathbf{I})\textbf{W}.
\end{equation}

When $\lambda$ is small, the Mahalanobis distance takes into account the correlations across dimensions.
However, it may overfit to the training set, since the metric matrix (\emph{i.e.} $WW^T$) is learnt from the training set which is usually small in person re-identification.
On the other hand, when $\lambda$ is large, the matrix $WW^T$ becomes close to the identity matrix.
In the extreme case, $WW^T$ equals to the identity matrix, and the distance
reduces to the Euclidean distance.
In this situation, the Euclidean distance does not consider the correlation, but may generalize robustly to unseen test sets.
So, we incorporate the advantage of the Mahalanobis and Euclidean distances and balance the matching accuracy and generalization performance via the constraint.


\section{Experiments}
\label{section_Experiments}

Our method is implemented via remodifying the CUDA-Convnet~\cite{krizhevsky2012imagenet} framework.
We report the evaluation with the one-shot standard protocol on three common benchmarks of person re-identification, \emph{i.e.} CUHK03~\cite{li2014deepreid}, CUHK01~\cite{li2012human} and VIPeR~\cite{gray2007evaluating}.

We begin with the description of CNN architecture we used for extracting features.
Then we report the evaluation on the validation set of CUHK03 for analyzing the effects of the moderate positive mining (Section~\ref{section_Analysis of Moderate Positive Mining}), the weight constraint (Section~\ref{section_Analysis of Weight Constraint}), and the CNN architecture (Section~\ref{section_Analysis of Untied Branches}).
Then, we compare our performance with the state-of-the-art methods on CUHK03 and CUHK01 (Section~\ref{section_Performance on CUHK03} and Section~\ref{section_Performance on CUHK01}).
Finally, we show the proposed method also performs well on the small data-set of VIPeR~\cite{gray2007evaluating} and gains competitive results (Section~\ref{section_Performance on VIPeR}).

\subsection{CNN architecture}
\label{section_CNN architecture}

\begin{figure}[!htb]
  \centering
  \includegraphics[width=0.56\textwidth]{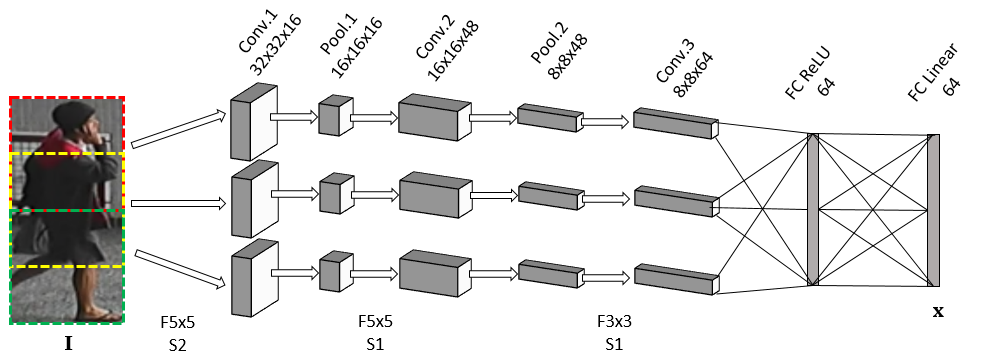}
  \caption{The CNN architecture. Top: layer type and output size. Bottom: convolution parameters with "F" and "S" denoting the filter size and stride, respectively.}
  \label{fig_CNN}
\end{figure}

The CNN is built by 3 branches with the details shown in Fig.~\ref{fig_CNN}.
The input image is normalized to $128\times64$ RGB. Then, it is split into three $64\times64$ overlapping color patches, each of which is charged by a branch. Each branch is constituted of 3 convolutional layers and 2 pooling layers. No parameter sharing is performed between branches. Then, the 3 branches are concluded by a FC layer with the ReLU activation. Finally, the output feature vector $\mathbf{x}$ is computed by another FC layer with linear activation.
For the computational stability, the features are normalized before sending to the metric learning layers.
The CNN and the metric layers are learned jointly via backward propagation.

Our network has much lighter weights (0.84M parameters) compared with the previous best methods on CUHK03\&01 (IDLA~\cite{ahmed2015improved}, 2.32M) and VIPeR (DeepFeature~\cite{ding2015deep}, 26M).
The reason that we build the CNN architecture in branches is to learn specific features from the different human body parts of pedestrian image; meanwhile, the morphological information is preserved from each part of human body.
DML~\cite{yi2014deep} adopted a similar architecture but with tied weights between branches.
In Section~\ref{section_Analysis of Untied Branches}, the experiments show the advantage of our untied architecture.

\subsection{Analysis of Moderate Positive Mining}
\label{section_Analysis of Moderate Positive Mining}

CUHK03 contains 1,369 subjects, each of which has around 10 images.
The default protocol randomly selects 1,169 subjects for training, 100 for validation, and 100 for test.
We pre-train the CNN with a softmax classification on the training set as the baseline.
The outputs of softmax correspond to the identities.

\begin{figure*}[!h]
  \centering
  \subfigure[]{
  \includegraphics[width=0.31\textwidth]{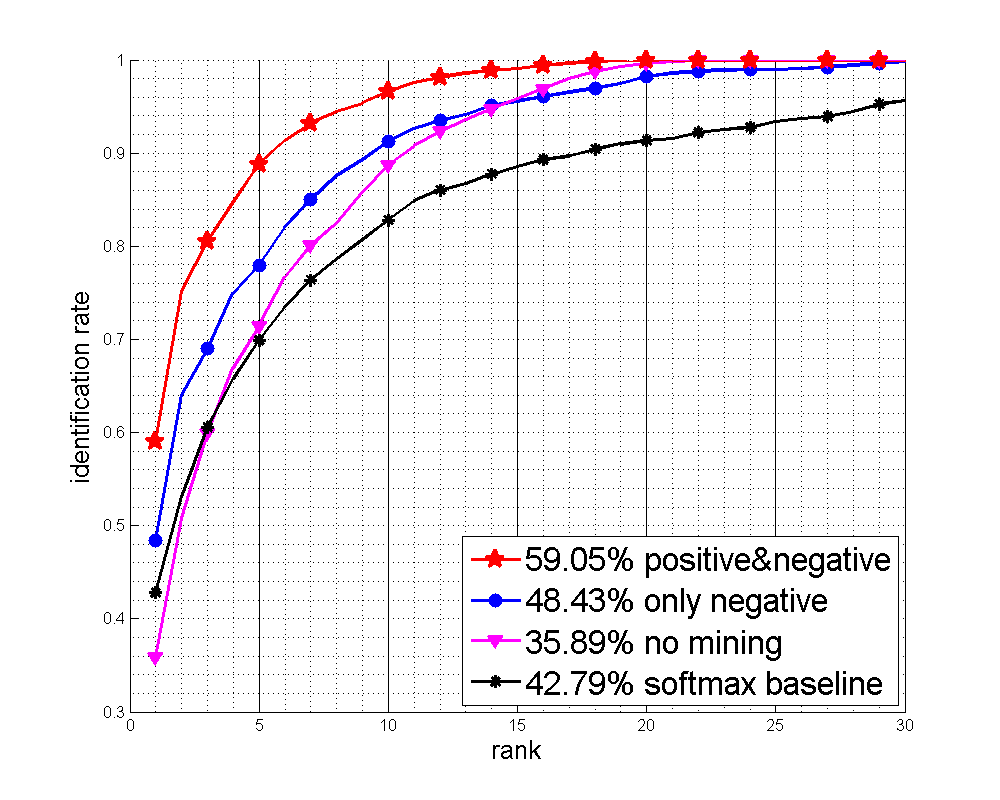}}
  \hspace{2pt}
  \subfigure[]{
  \includegraphics[width=0.29\textwidth]{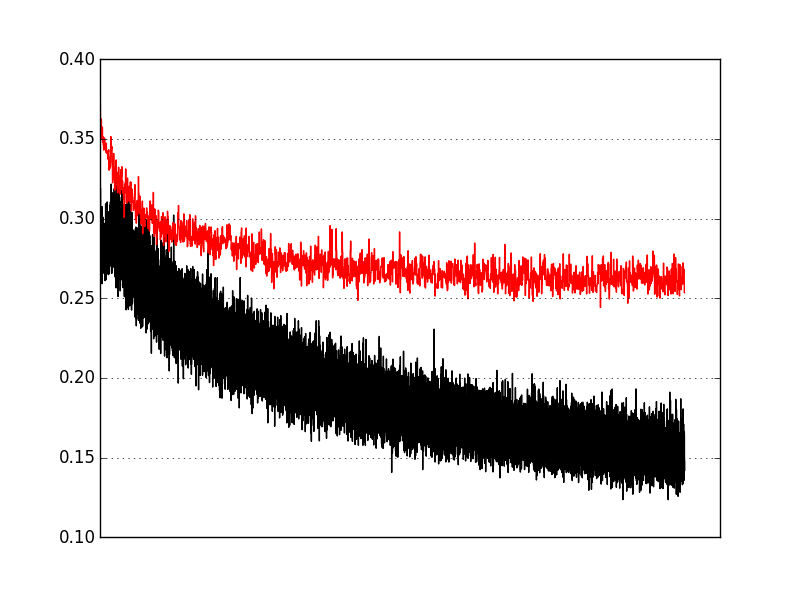}}
  \subfigure[]{
  \includegraphics[width=0.33\textwidth]{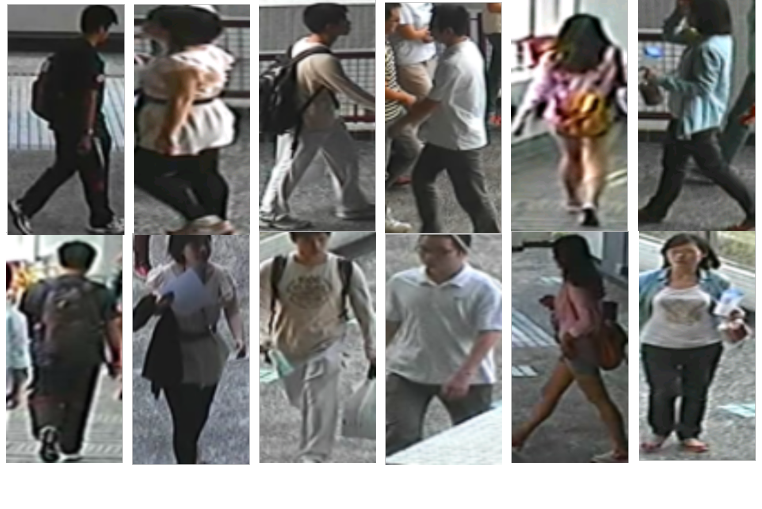}}
  \caption{(a) Performance analysis of moderate positive mining.
  Red: both moderate positive mining and hard negative mining are employed.
  Blue: only hard negative mining is employed.
  Magenta: no mining technique is employed during training.
  Black: the softmax baseline.
  (b) The loss curves along training iterations. Black: training set. Red: validation set.
  (c) Some positives mined by the moderate positive mining method.}

  \label{fig_mpm_compare}
\end{figure*}

To demonstrate the advantage of moderate positive mining, we compare the performances on the validation set with and without the moderate positive mining.
The cumulative matching characteristic (CMC) curves and the rank-1 identification rates are shown in Fig.~\ref{fig_mpm_compare}(a).
We can find that the collaboration of moderate positive mining and hard negative mining achieves the best result (red line).
The absence of moderate positive mining leads to a significant derogation of performance (blue).
This reflects that the manifold is badly learned if all the positive pairs are used undiscriminatingly.

If both of the two mining methods are not used (magenta), the network gives very low identification rate at low ranks, even worse than the baseline (black).
This indicates that moderate positive mining and hard negative mining are both crucial for training.

The CMC curves of the 3 trained networks tend to be saturated after the rank exceeds 20, whereas the baseline network remains at a relatively low identification rate.
This indicates that the training with the metric layers is the basic contributor of the improvement.

The training of network converges well as the loss value descending with respect to the iterations (shown in Fig.~\ref{fig_mpm_compare}(b)). Some positives, which are mined by moderate positive mining during training, are shown in Fig.~\ref{fig_mpm_compare}(c). These positives are with moderate extent of difficulty compared with those hard ones in Fig.~\ref{fig_hard-positives}(a).

\subsection{Analysis of Weight Constraint}
\label{section_Analysis of Weight Constraint}

\begin{figure*}[!h]
  \centering
  \subfigure[]{
  \includegraphics[width=0.46\textwidth]{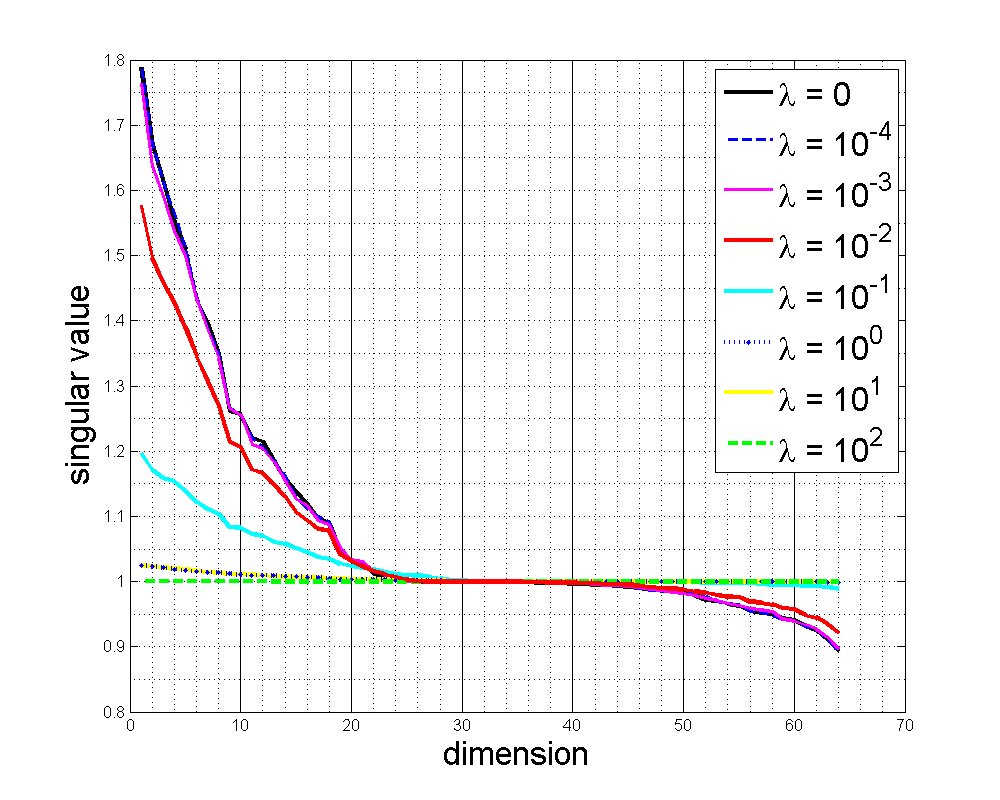}}
  \hspace{2pt}
  \subfigure[]{
  \includegraphics[width=0.46\textwidth]{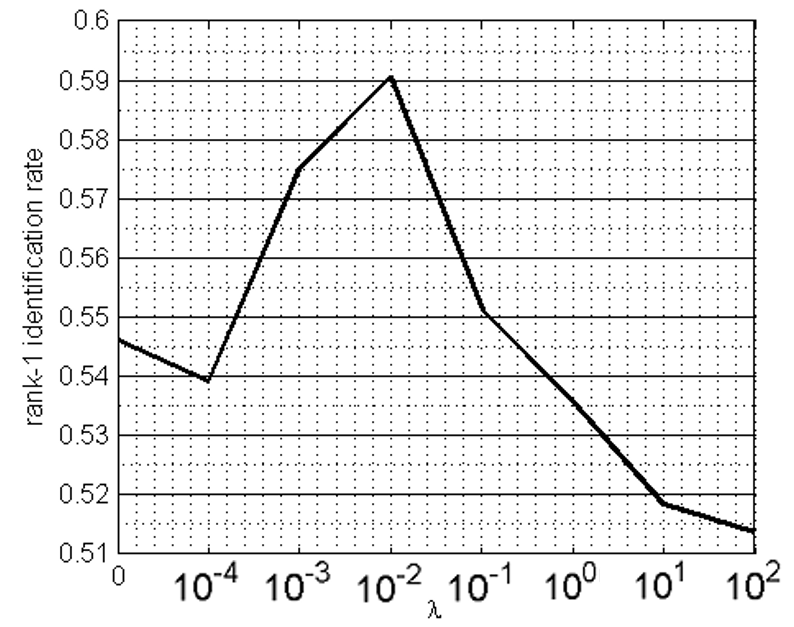}}

  \caption{(a) The spectrums of the matrix $\textbf{M}$. The spectrums with $\lambda=10^1, 10^0$ are very close; those with $\lambda=10^{-3}, 10^{-4}, 0$ are also very close. Best viewed in color.
  (b) The rank-1 identification rates with different $\lambda$ of the weight constraint. }
  \label{fig_lambda}
\end{figure*}

We inspect the metric matrices learned with different relative weights ($\lambda$) of the regularization.
In Fig.~\ref{fig_lambda}(a), we show the spectrums of the matrix $\textbf{M}$.
We also show the corresponding rank-1 identification rates in Fig.~\ref{fig_lambda}(b).

When $\lambda = 10^2$, the singular values are almost constant at 1, which means the metric layers almost give the Euclidean distance. This leads to the low variance and high bias.
As $\lambda$ increases, the matrix has varying singular values across dimensions.
This implies that the learned metric suits the training data well, but is more likely to have over-fitting.
Therefore, a moderate value of $\lambda$ gives a trade-off between the variance and bias, which is an appropriate choice for good performance (Fig.~\ref{fig_lambda}(b)).

\subsection{Analysis of Untied Branches}
\label{section_Analysis of Untied Branches}

\begin{figure*}[!h]
  \centering
  \subfigure[]{
  \includegraphics[width=0.46\textwidth]{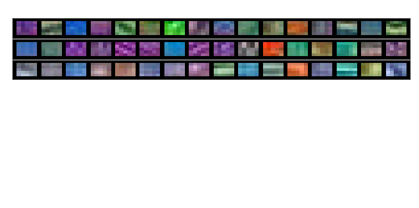}}
  \hspace{2pt}
  \subfigure[]{
  \includegraphics[width=0.46\textwidth]{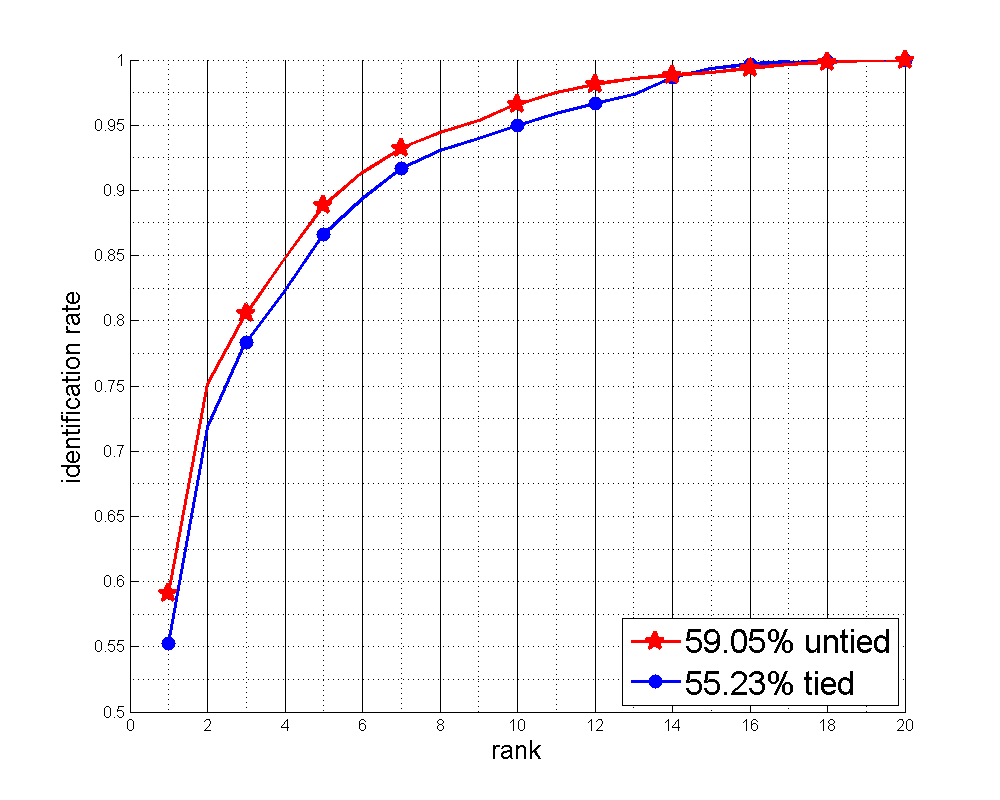}}

  \caption{(a) The learned filters of the first convolutional layer. The top, middle and bottom lines correspond to the 3 branches in the proposed CNN. Best viewed in color.
  (b) The performances with and without tied weights between branches.}
  \label{fig_branch}
\end{figure*}

We show the learned filters of untied branches in Fig.~\ref{fig_branch}(a).
The network has learned remarkable color representations, which is coherent with the results of IDLA~\cite{ahmed2015improved}.
Since we apply untied weights between branches, each branch learns different filters from their own part.
As shown in Fig.~\ref{fig_branch}(a) where each row demonstrates a filter set from one branch,
we can find each branch has its own emphasis in color.
For example, the middle branch inclines to violet and blue, whereas the bottom branch has learned filters of obviously lighter colors than the other two.
The reason is that pedestrian images have regular appearance of human body.
Each part has its own color distribution.
Therefore, the branches learn the part-specific filters, the morphological information is taken into account for the features.

We compare the performances with and without tied weights between branches in Fig.~\ref{fig_branch}(b). We augment the filter number in the tied-branches network so to make roughly equal parameter number with the untied-branch.
The untied-branch network gains a better performance than that of tied branches.
It reflects that, when the network has a certain complexity, the neural structure (\emph{i.e.} tied vs untied) becomes very important.
How to organize the network structure is a critical issue for good performance.

\subsection{Performance on CUHK03}
\label{section_Performance on CUHK03}
We adopt a random translation for the training data augmentation.
The images are randomly cropped (0-5 pixels) in horizon and vertical, and stretched to recover the size.
According to the validation results (Section~\ref{section_Analysis of Weight Constraint}), we set the parameter $\lambda = 10^{-2}$ in all the following experiments.
The moderate positive mining and hard negative mining are employed.

\begin{figure*}[!htbp]
  \centering
  \subfigure[labeled]{
  \includegraphics[width=0.46\textwidth]{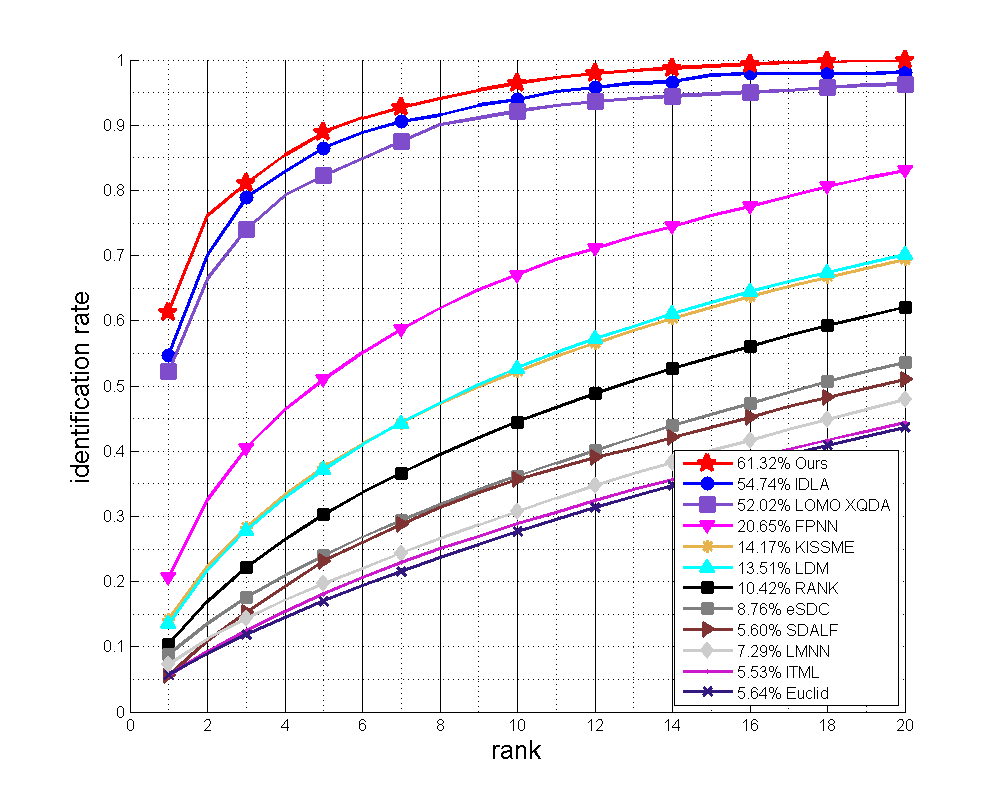}}
  \subfigure[detected]{
  \includegraphics[width=0.46\textwidth]{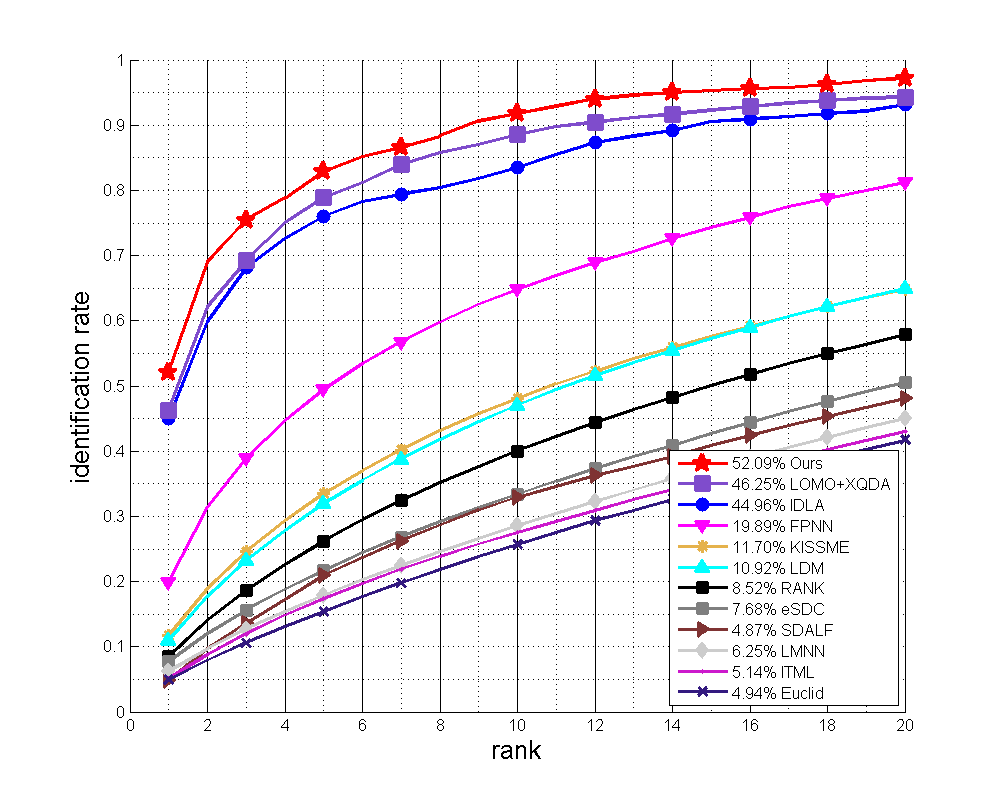}}

  \caption{CMC curves and rank-1 identification rates on the CUHK03 data set. Our method outperforms the previous methods on both labeled (a) and detected (b) versions.}
  \label{fig_cmc_cuhk03}
\end{figure*}

CUHK03 has 2 versions, one has manually labeled images, and the other has detected images. We evaluate our method on the test set of both versions.
We compare our performance with the traditional methods and deep learning methods. The traditional methods include LOMO-XQDA~\cite{liao2015person}, KISSME~\cite{koestinger2012large}, LDM~\cite{guillaumin2009you}, RANK~\cite{mcfee2010metric}, eSDC~\cite{zhao2013unsupervised}, SDALF~\cite{farenzena2010person}, LMNN~\cite{weinberger2005distance}, ITML~\cite{davis2007information}, Euclid~\cite{zhao2013unsupervised}. The deep learning methods include FPNN~\cite{li2014deepreid} and IDLA~\cite{ahmed2015improved}. IDLA and LOMO-XQDA gained the previously best performance on CUHK03.
The CMC curves and the rank-1 identification rates are shown in Fig.~\ref{fig_cmc_cuhk03}.
Our method achieves better performance than the previous state-of-the-art methods on not only the labeled version but also the detected version.
This indicates that our method achieves good robustness to the misalignment of detection.

\subsection{Performance on CUHK01}
\label{section_Performance on CUHK01}

\begin{figure*}[!htbp]
  \centering
  \subfigure[CUHK01]{
  \includegraphics[width=0.46\textwidth]{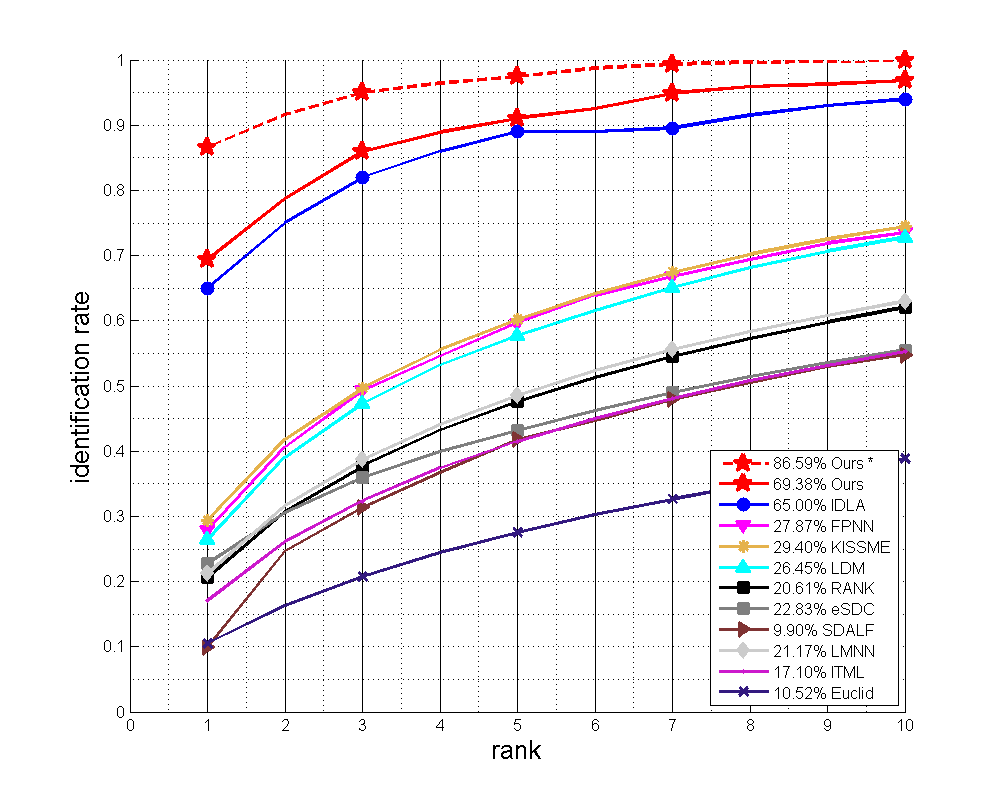}}
  \subfigure[VIPeR]{
  \includegraphics[width=0.46\textwidth]{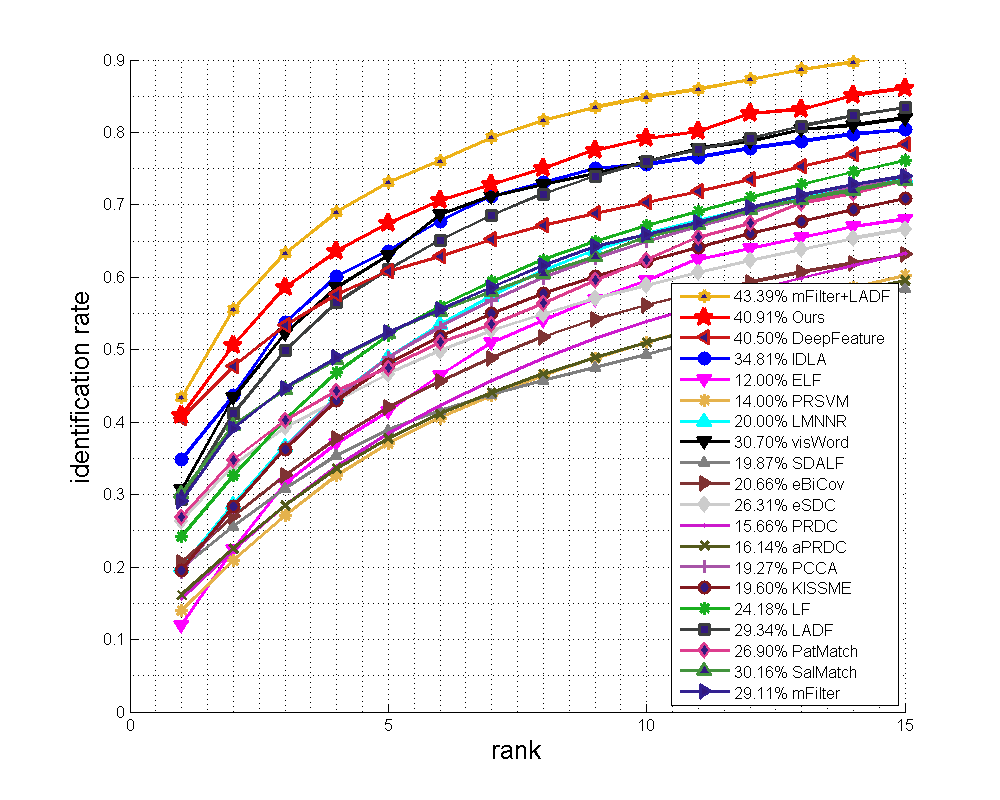}}

  \caption{CMC curves and rank-1 identification rates on CUHK01 (a) and VIPeR (b).}
  \label{fig_cmc_cuhk01_viper}
\end{figure*}

The CUHK01 data set contains 971 subjects, each of which has 4 images under 2 camera views.
According to the protocol in~\cite{li2012human}, the data set is divided into a training set of 871 subjects and a test set of 100.
We train the network on CUHK03, and further fine-tune it on CUHK01, as the same setting with the state-of-the-art method IDLA~\cite{ahmed2015improved}.
We compare our approach with the previously mentioned methods. The CMC curves and rank-1 identification rates are shown in Fig.~\ref{fig_cmc_cuhk01_viper}(a).
Our approach gains the best result (the red line) with 69\% rank-1 identification rate.

\begin{figure*}[!htb]
  \centering
  \includegraphics[width=0.9\textwidth]{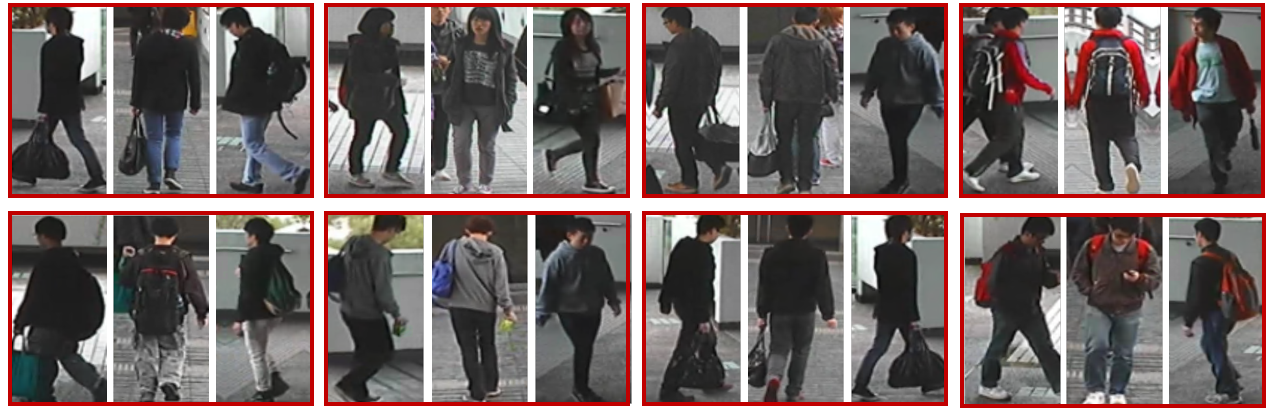}
  \caption{Some failed cases on CUHK01 by the proposed method. Left: true gallery. Middle: probe. Right: false positive.}
  \label{fig_cuhk01_failed}
\end{figure*}

Besides, to inspect the limitation of the data set CUHK01, we involve the recently released Market1501~\cite{zheng2015scalable} into the training. As the training data increases, our network gives a better performance (the red dash line marked as ``\emph{Ours *}'') with 87\% rank-1 identification rate.
We show certain failed cases in Fig.~\ref{fig_cuhk01_failed}. In each block, we give the true gallery, probe and false positive image from left to right.
We find that most failed cases come from the dark color images or the negative pairs with significant color correspondence.
This phenomenon is in line with the fact~\cite{ahmed2015improved} that the learned filters in network mainly focus on image colors (as shown in Fig.~\ref{fig_branch}(a)).
The re-identification problem becomes extremely difficult when the true positive pairs have inconsistent colors in view while the negative pairs have similar colors (due to the lighting, camera setting \emph{etc.}).

\subsection{Performance on VIPeR}
\label{section_Performance on VIPeR}

The VIPeR~\cite{gray2007evaluating} data set includes 632 subjects, each of which has 2 images from two different cameras.
Although VIPeR is a small data set which is not suitable for training CNN, we are still interested in the performance on this challenging task.
The data set is randomly split into two subsets, each has non-overlapping subjects of the same size. The two subsets are for either training or test.
We fine-tune the network on the 316-person training set and test it on the test set.
We also adopt a random translation for training data augmentation.
The results are shown in Fig.~\ref{fig_cmc_cuhk01_viper}(b).
We compare our model with IDLA~\cite{ahmed2015improved}, DeepFeature~\cite{ding2015deep}, visual word (visWord)~\cite{zhang2014novel}, saliency matching (SalMatch), patch matching (PatMatch)~\cite{zhao2013person}, ELF~\cite{gheissari2006person}, PRSVM~\cite{bazzani2012multiple}, LMNNR~\cite{bak2011multiple}, eBiCov~\cite{ma2012bicov}, local Fisher discriminant analysis (LF)~\cite{pedagadi2013local}, PRDC~\cite{zheng2011person}, aPRDC~\cite{liu2012person}, PCCA~\cite{mignon2012pcca}, mid-level filters (mFilter)~\cite{zhao2014learning} and the fusion of mFilter and LADF~\cite{li2013learning}.
Our approach achieves the identification rate of 40.91\% at rank 1, which is the best result on VIPeR compared with the existing deep learning methods.
Note that the highest rank-1 identification rate (43.39\%) is obtained by a combination of two methods (mFilter+LADF)~\cite{li2013learning}.
The identification rate by DeepFeature~\cite{ding2015deep} is close to ours at rank 1, but much lower at higher ranks.

\section{Conclusion}
\label{section_Conclusion}

The large variations of pedestrian data is a challenging point for the person re-identification methods.
Although CNN has a strong ability to extract features, pedestrian data follows the very irregular distribution in the feature space due to the large variations.
In order to cope with the problem and train the robust deep embedding, the positive training samples should be selected deliberately.
In this paper, we propose a novel moderate positive mining method to embed robust deep metric for person re-identification.
We find that mining the moderate positive samples is crucial for training deep networks, especially when it comes to the difficult data with large intra-class variations (\emph{e.g.} pedestrian).
The moderate positive mining method dynamically select the suitable positive pairs for learning robust embedding adaptive to the data manifold.
Moreover, we propose the weight constraint for gaining the good robustness to the over-fitting problem in person re-identification.

Due to these improvements, our method achieves state-of-the-art performances on CUHK03 and CUHK01, and competitive results on VIPeR.
By mining the moderate positive samples for the training, we can reduce the intra-class variance while preserving the intrinsic graphical structure of pedestrian data;
the metric weight constraint helps to improve the generalization ability of the network, especially when the most parameters are in the metric layers.

\section{Acknowledgement}
\label{section_Acknowledgement}
This work was supported by the National Key Research and Development Plan (Grant No.2016YFC0801003), the Chinese National Natural Science Foundation Projects \#61473291, \#61572501, \#61502491, \#61572536, NVIDIA GPU donation program and AuthenMetric R\&D Funds.

\clearpage
\bibliographystyle{splncs03}
\bibliography{eccvbib}

\begin{thebibliography}{10}
\providecommand{\url}[1]{\texttt{#1}}
\providecommand{\urlprefix}{URL }

\bibitem{ahmed2015improved}
Ahmed, E., Jones, M., Marks, T.K.: An improved deep learning architecture for
  person re-identification. In: Computer Vision and Pattern Recognition (CVPR),
  2015 IEEE Conference on. IEEE (2015)

\bibitem{bak2011multiple}
Bak, S., Corvee, E., Bremond, F., Thonnat, M.: Multiple-shot human
  re-identification by mean riemannian covariance grid. In: Advanced Video and
  Signal-Based Surveillance (AVSS), 2011 8th IEEE International Conference on.
  pp. 179--184. IEEE (2011)

\bibitem{bazzani2012multiple}
Bazzani, L., Cristani, M., Perina, A., Murino, V.: Multiple-shot person
  re-identification by chromatic and epitomic analyses. Pattern Recognition
  Letters  33(7),  898--903 (2012)

\bibitem{belkin2003laplacian}
Belkin, M., Niyogi, P.: Laplacian eigenmaps for dimensionality reduction and
  data representation. Neural computation  15(6),  1373--1396 (2003)

\bibitem{davis2007information}
Davis, J.V., Kulis, B., Jain, P., Sra, S., Dhillon, I.S.: Information-theoretic
  metric learning. In: Proceedings of the 24th international conference on
  Machine learning. pp. 209--216. ACM (2007)

\bibitem{ding2015deep}
Ding, S., Lin, L., Wang, G., Chao, H.: Deep feature learning with relative
  distance comparison for person re-identification. Pattern Recognition  (2015)

\bibitem{farenzena2010person}
Farenzena, M., Bazzani, L., Perina, A., Murino, V., Cristani, M.: Person
  re-identification by symmetry-driven accumulation of local features. In:
  Computer Vision and Pattern Recognition (CVPR), 2010 IEEE Conference on. pp.
  2360--2367. IEEE (2010)

\bibitem{gheissari2006person}
Gheissari, N., Sebastian, T.B., Hartley, R.: Person reidentification using
  spatiotemporal appearance. In: Computer Vision and Pattern Recognition, 2006
  IEEE Computer Society Conference on. vol.~2, pp. 1528--1535. IEEE (2006)

\bibitem{gray2007evaluating}
Gray, D., Brennan, S., Tao, H.: Evaluating appearance models for recognition,
  reacquisition, and tracking. In: Proc. IEEE International Workshop on
  Performance Evaluation for Tracking and Surveillance (PETS). vol.~3. Citeseer
  (2007)

\bibitem{guillaumin2009you}
Guillaumin, M., Verbeek, J., Schmid, C.: Is that you? metric learning
  approaches for face identification. In: Computer Vision, 2009 IEEE 12th
  International Conference on. pp. 498--505. IEEE (2009)

\bibitem{hu2014discriminative}
Hu, J., Lu, J., Tan, Y.P.: Discriminative deep metric learning for face
  verification in the wild. In: Computer Vision and Pattern Recognition (CVPR),
  2014 IEEE Conference on. pp. 1875--1882. IEEE (2014)

\bibitem{khamis2014joint}
Khamis, S., Kuo, C.H., Singh, V.K., Shet, V.D., Davis, L.S.: Joint learning for
  attribute-consistent person re-identification. In: Computer Vision-ECCV 2014
  Workshops. pp. 134--146. Springer (2014)

\bibitem{koestinger2012large}
Koestinger, M., Hirzer, M., Wohlhart, P., Roth, P.M., Bischof, H.: Large scale
  metric learning from equivalence constraints. In: Computer Vision and Pattern
  Recognition (CVPR), 2012 IEEE Conference on. pp. 2288--2295. IEEE (2012)

\bibitem{krizhevsky2012imagenet}
Krizhevsky, A., Sutskever, I., Hinton, G.E.: Imagenet classification with deep
  convolutional neural networks. In: Advances in neural information processing
  systems. pp. 1097--1105 (2012)

\bibitem{li2013locally}
Li, W., Wang, X.: Locally aligned feature transforms across views. In: Computer
  Vision and Pattern Recognition (CVPR), 2013 IEEE Conference on. pp.
  3594--3601. IEEE (2013)

\bibitem{li2012human}
Li, W., Zhao, R., Wang, X.: Human reidentification with transferred metric
  learning. In: ACCV (1). pp. 31--44 (2012)

\bibitem{li2014deepreid}
Li, W., Zhao, R., Xiao, T., Wang, X.: Deepreid: Deep filter pairing neural
  network for person re-identification. In: Computer Vision and Pattern
  Recognition (CVPR), 2014 IEEE Conference on. pp. 152--159. IEEE (2014)

\bibitem{li2013learning}
Li, Z., Chang, S., Liang, F., Huang, T.S., Cao, L., Smith, J.R.: Learning
  locally-adaptive decision functions for person verification. In: Computer
  Vision and Pattern Recognition (CVPR), 2013 IEEE Conference on. pp.
  3610--3617. IEEE (2013)

\bibitem{liao2015person}
Liao, S., Hu, Y., Zhu, X., Li, S.Z.: Person re-identification by local maximal
  occurrence representation and metric learning. In: Proceedings of the IEEE
  Conference on Computer Vision and Pattern Recognition. pp. 2197--2206 (2015)

\bibitem{liu2012person}
Liu, C., Gong, S., Loy, C.C., Lin, X.: Person re-identification: What features
  are important? In: Computer Vision--ECCV 2012. Workshops and Demonstrations.
  pp. 391--401. Springer (2012)

\bibitem{ma2012bicov}
Ma, B., Su, Y., Jurie, F.: Bicov: a novel image representation for person
  re-identification and face verification. In: British Machive Vision
  Conference. pp. 11--pages (2012)

\bibitem{manly2004multivariate}
Manly, B.F.: Multivariate statistical methods: a primer. CRC Press (2004)

\bibitem{martinel2014saliency}
Martinel, N., Micheloni, C., Foresti, G.L.: Saliency weighted features for
  person re-identification. In: Computer Vision-ECCV 2014 Workshops. pp.
  191--208. Springer (2014)

\bibitem{mcfee2010metric}
McFee, B., Lanckriet, G.R.: Metric learning to rank. In: Proceedings of the
  27th International Conference on Machine Learning (ICML-10). pp. 775--782
  (2010)

\bibitem{mignon2012pcca}
Mignon, A., Jurie, F.: Pcca: A new approach for distance learning from sparse
  pairwise constraints. In: Computer Vision and Pattern Recognition (CVPR),
  2012 IEEE Conference on. pp. 2666--2672. IEEE (2012)

\bibitem{paisitkriangkrai2015learning}
Paisitkriangkrai, S., Shen, C., van~den Hengel, A.: Learning to rank in person
  re-identification with metric ensembles. In: Proceedings of the IEEE
  Conference on Computer Vision and Pattern Recognition. pp. 1846--1855 (2015)

\bibitem{parkhi2015deep}
Parkhi, O.M., Vedaldi, A., Zisserman, A.: Deep face recognition. Proceedings of
  the British Machine Vision  (2015)

\bibitem{pedagadi2013local}
Pedagadi, S., Orwell, J., Velastin, S., Boghossian, B.: Local fisher
  discriminant analysis for pedestrian re-identification. In: Computer Vision
  and Pattern Recognition (CVPR), 2013 IEEE Conference on. pp. 3318--3325. IEEE
  (2013)

\bibitem{roweis2000nonlinear}
Roweis, S.T., Saul, L.K.: Nonlinear dimensionality reduction by locally linear
  embedding. Science  290(5500),  2323--2326 (2000)

\bibitem{schroff2015facenet}
Schroff, F., Kalenichenko, D., Philbin, J.: Facenet: A unified embedding for
  face recognition and clustering. In: Proceedings of the IEEE Conference on
  Computer Vision and Pattern Recognition (2015)

\bibitem{tenenbaum2000global}
Tenenbaum, J.B., De~Silva, V., Langford, J.C.: A global geometric framework for
  nonlinear dimensionality reduction. science  290(5500),  2319--2323 (2000)

\bibitem{weinberger2005distance}
Weinberger, K.Q., Blitzer, J., Saul, L.K.: Distance metric learning for large
  margin nearest neighbor classification. In: Advances in neural information
  processing systems. pp. 1473--1480 (2005)

\bibitem{xiong2014person}
Xiong, F., Gou, M., Camps, O., Sznaier, M.: Person re-identification using
  kernel-based metric learning methods. In: Computer Vision--ECCV 2014, pp.
  1--16. Springer (2014)

\bibitem{yang2014salient}
Yang, Y., Yang, J., Yan, J., Liao, S., Yi, D., Li, S.Z.: Salient color names
  for person re-identification. In: Computer Vision--ECCV 2014, pp. 536--551.
  Springer (2014)

\bibitem{yi2014deep}
Yi, D., Lei, Z., Li, S.Z.: Deep metric learning for practical person
  re-identification. arXiv preprint arXiv:1407.4979  (2014)

\bibitem{zhang2014prism}
Zhang, Z., Saligrama, V.: Prism: Person re-identification via structured
  matching. arxiv preprint. IEEE Transaction on Pattern Analysis and Machine
  Intelligence  (2015)

\bibitem{zhang2014novel}
Zhang, Z., Chen, Y., Saligrama, V.: A novel visual word co-occurrence model for
  person re-identification. In: Computer Vision-ECCV 2014 Workshops. pp.
  122--133. Springer (2014)

\bibitem{zhang2015group}
Zhang, Z., Chen, Y., Saligrama, V.: Group membership prediction. In: Computer
  Vision (ICCV), 2015 IEEE International Conference on. IEEE (2015)

\bibitem{zhao2013person}
Zhao, R., Ouyang, W., Wang, X.: Person re-identification by salience matching.
  In: Computer Vision (ICCV), 2013 IEEE International Conference on. pp.
  2528--2535. IEEE (2013)

\bibitem{zhao2013unsupervised}
Zhao, R., Ouyang, W., Wang, X.: Unsupervised salience learning for person
  re-identification. In: Computer Vision and Pattern Recognition (CVPR), 2013
  IEEE Conference on. pp. 3586--3593. IEEE (2013)

\bibitem{zhao2014learning}
Zhao, R., Ouyang, W., Wang, X.: Learning mid-level filters for person
  re-identification. In: Computer Vision and Pattern Recognition (CVPR), 2014
  IEEE Conference on. pp. 144--151. IEEE (2014)

\bibitem{zheng2015scalable}
Zheng, L., Shen, L., Tian, L., Wang, S., Wang, J., Tian, Q.: Scalable person
  re-identification: A benchmark. In: Computer Vision, IEEE International
  Conference on (2015)

\bibitem{zheng2011person}
Zheng, W.S., Gong, S., Xiang, T.: Person re-identification by probabilistic
  relative distance comparison. In: Computer Vision and Pattern Recognition
  (CVPR), 2011 IEEE Conference on. pp. 649--656. IEEE (2011)

\end{thebibliography}
\end{document}